\documentclass[conference]{IEEEtran}
\IEEEoverridecommandlockouts
\usepackage{cite}
\usepackage{amsmath,amssymb,amsfonts}
\usepackage{algorithmic}
\usepackage{graphicx}
\usepackage{textcomp}
\usepackage{caption}
\usepackage{subcaption}
\usepackage{hyperref}
\usepackage[T1]{fontenc}

\def\BibTeX{{\rm B\kern-.05em{\sc i\kern-.025em b}\kern-.08em
    T\kern-.1667em\lower.7ex\hbox{E}\kern-.125emX}}



\begin{document}

\title{Towards an Autonomous Surface Vehicle Prototype for Artificial Intelligence Applications of Water Quality Monitoring\\

\thanks{This work was partially funded by the following projects:  TED2021-131326B-C21 of the Spanish Ministry of Science MCIN/ AEI/ 10.13039/ 501100011033, and the European Union with "NextGenerationEU / PRTR" funds,
TED2021-131326A-C22 of the Spanish Ministry of Science MCIN/ AEI/ 10.13039/ 501100011033 and the European Union with "Next Generation EU / PRTR" funds, and the "Junta de Andalucía: Consejería de Universidad, Investigación e Innovación" through the project "Monitorization of Environmental Dangers with Unmanned Surface Agents: (MEDUSA)" under Grant PCM\_00019, public funding. 
Activity: C23.I1.P03.S01.01 Andalucía Public subsidy for the development of the «INVESTIGO Programme», financed by funds from the «Recovery and Resilience Mechanism», Regional Government of Andalucía and the European Union}
}

\author{\IEEEauthorblockN{1\textsuperscript{st} Luis Miguel Díaz}
\IEEEauthorblockA{\textit{Electronic Engineering Department} \\
\textit{University of Sevilla}\\
Sevilla, Spain \\
luismidp7@gmail.com}
\and
\IEEEauthorblockN{2\textsuperscript{nd} Samuel Yanes Luis}
\IEEEauthorblockA{\textit{Electronic Engineering Department} \\
\textit{University of Sevilla}\\
Sevilla, Spain \\
syanes@us.es}
\and
\IEEEauthorblockN{3\textsuperscript{rd} Alejandro Mendoza Barrionuevo}
\IEEEauthorblockA{\textit{Electronic Engineering Department} \\
\textit{University of Sevilla}\\
Sevilla, Spain \\
amendoza1@us.es}
\and
\IEEEauthorblockN{4\textsuperscript{th} Dame Seck Diop}
\IEEEauthorblockA{\textit{Electronic Engineering Department} \\
\textit{University of Sevilla}\\
Sevilla, Spain \\
dseck@us.es}
\and
\IEEEauthorblockN{5\textsuperscript{th} Manuel Perales}
\IEEEauthorblockA{\textit{Electronic Engineering Department} \\
\textit{University of Sevilla}\\
Sevilla, Spain \\
mperales@us.es}
\and
\IEEEauthorblockN{6\textsuperscript{th} Alejandro Casado}
\IEEEauthorblockA{\textit{Electronic Engineering Department} \\
\textit{University of Sevilla}\\
Sevilla, Spain \\
alexcpesp@gmail.com}
\and
\IEEEauthorblockN{7\textsuperscript{th} Sergio Toral}
\IEEEauthorblockA{\textit{Electronic Engineering Department} \\
\textit{University of Sevilla}\\
Sevilla, Spain \\
storal@us.es}
\and
\IEEEauthorblockN{8\textsuperscript{th} Daniel Gutiérrez}
\IEEEauthorblockA{\textit{Electronic Engineering Department} \\
\textit{University of Sevilla}\\
Sevilla, Spain \\
dgutierrezreina@us.es}
}

\maketitle

\begin{abstract}

The use of Autonomous Surface Vehicles, equipped with water quality sensors and artificial vision systems, allows for a smart and adaptive deployment in water resources environmental monitoring. This paper presents a real implementation of a vehicle prototype that to address the use of Artificial Intelligence algorithms and enhanced sensing techniques for water quality monitoring. The vehicle is fully equipped with high-quality sensors to measure water quality parameters and water depth. Furthermore, by means of a stereo-camera, it also can detect and locate macro-plastics in real environments by means of deep visual models, such as YOLOv5. In this paper, experimental results, carried out in Lago Mayor (Sevilla), has been presented as proof of the capabilities of the proposed architecture. The overall system, and the early results obtained, are expected to provide a solid example of a real platform useful for the water resource monitoring task, and to serve as a real case scenario for deploying Artificial Intelligence algorithms, such as path planning, artificial vision, etc.

\end{abstract}


\section{Introduction}


The urgent need to improve the efficiency of data collection in large aquatic environments, such as seaports, rivers, and lakes, has driven significant technological challenges: adaptive monitorization, large measurement campaigns, reactive path planning behavior, etc. These challenges, have pushed the development of innovative solutions in the field of robotics, like Autonomous Surface Vehicles (ASVs) \cite{survey_dl_asv}. ASVs offer several advantages, including being able to operate autonomously, reducing the need for constant human supervision, especially in aquatic environments that are challenging or dangerous for divers or human operators. Equipped with a variety of specialized sensors and equipment, ASVs can perform a wide range of tasks, such as oil spill detection \cite{oilcleaner}, water quality monitoring or underwater mapping, making them valuable tools for a variety of industries, including environmental management, maritime safety and scientific research. Advances in sensing and navigation technology have further improved the efficiency and accuracy of ASVs, enabling them to autonomously avoid obstacles using object detection systems \cite{ren2021autonomous, asv_object_detection}, increasing their safety and reliability in dynamic environments.

\begin{figure}[tbp]
    \centering
    \includegraphics[width=0.9\linewidth]{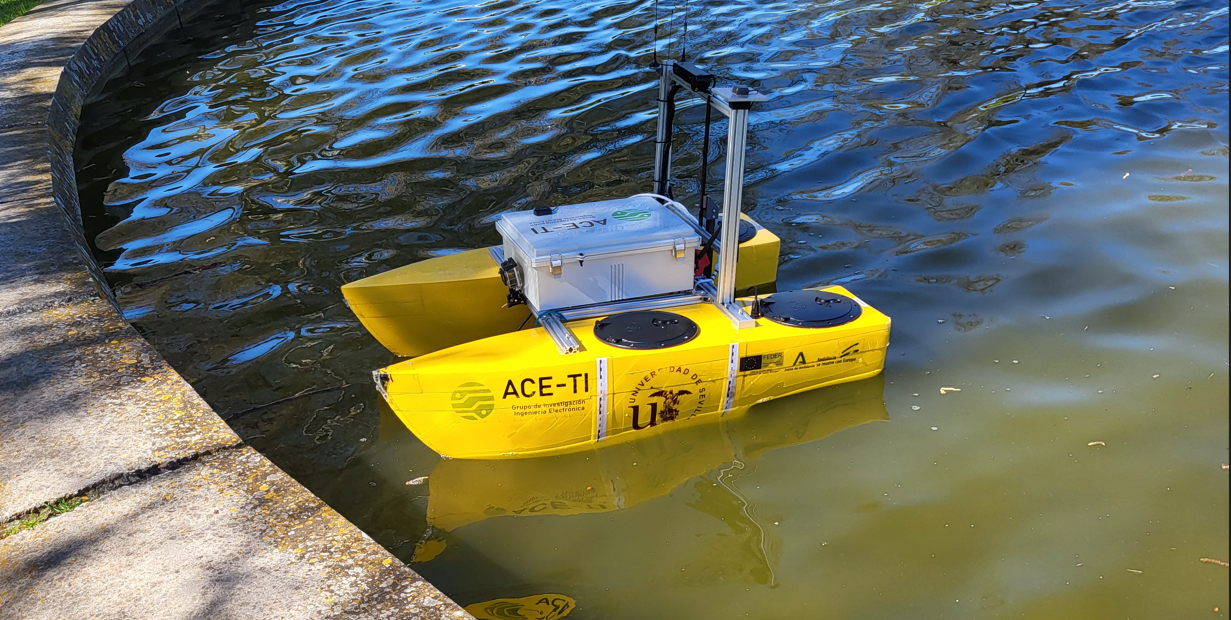}
    \caption{Photo of the ASV prototype deployed for WQP sampling in Lago Mayor, Parque del Alamillo, Sevilla.}
    \label{fig:drones_in_alamillo}
\end{figure}

Despite the aforementioned significant progress in the development and adoption of ASV technologies, significant challenges remain uncovered, such as the need for designing robust systems for large-scale environmental monitoring without direct supervision, as well as debris detection and cleanup systems. All these challenges are related to the real application of AI techniques and the problems related to a real deployment scenario. In this paper, a novel implementation of ASV is presented (see Fig. \ref{fig:drones_in_alamillo}), in which the hardware and software implementations are described to address the real challenges of a functional AI-equipped vehicle in a real aquatic environment. For the hardware, all the necessary commercial modules will be described: water quality sensors, navigation and computation modules, among others. For the software, an open-source software architecture based on ROS2 in proposed with an open-source versatile middleware for robotic applications. The software architecture tries to unify the different ROS2 nodes into an asynchronous and robust behavior, that can handle different AI techniques like object detection with YOLOv5 or Informative Path Planning optimization, with exterior communications as well. In addition, this paper provides an early proof of concept with water quality parameter data collected in test missions conducted in a lake. Future directions and open challenges will be also discussed, specifically the detection of macro-plastics in ports or seashores and other adaptive monitorization algorithms that can be easily deployed in this platform.

This paper is organized as follows: In Section \ref{Hardware architecture} are detailed the fundamental elements that compose the vehicle, from navigation systems to the specialized sensors that enable it to perform the designated tasks. Section \ref{Software architecture} describes the communication protocols used to facilitate the interaction between components and an overview of how the ASV is controlled and monitored during its operations. Section \ref{Experimental Results} presents the results obtained during an experimental mission at Lago Mayor, Parque del Alamillo (Sevilla). Finally, the Section \ref{Conclusion and Future Works} reflects on the achievements to date and outlines possible directions for continuous improvement and future development.

\section{Hardware architecture} 
\label{Hardware architecture}

\begin{figure}[t]
\centering
    \begin{subfigure}{0.45\linewidth}
        \includegraphics[width=0.9\linewidth]{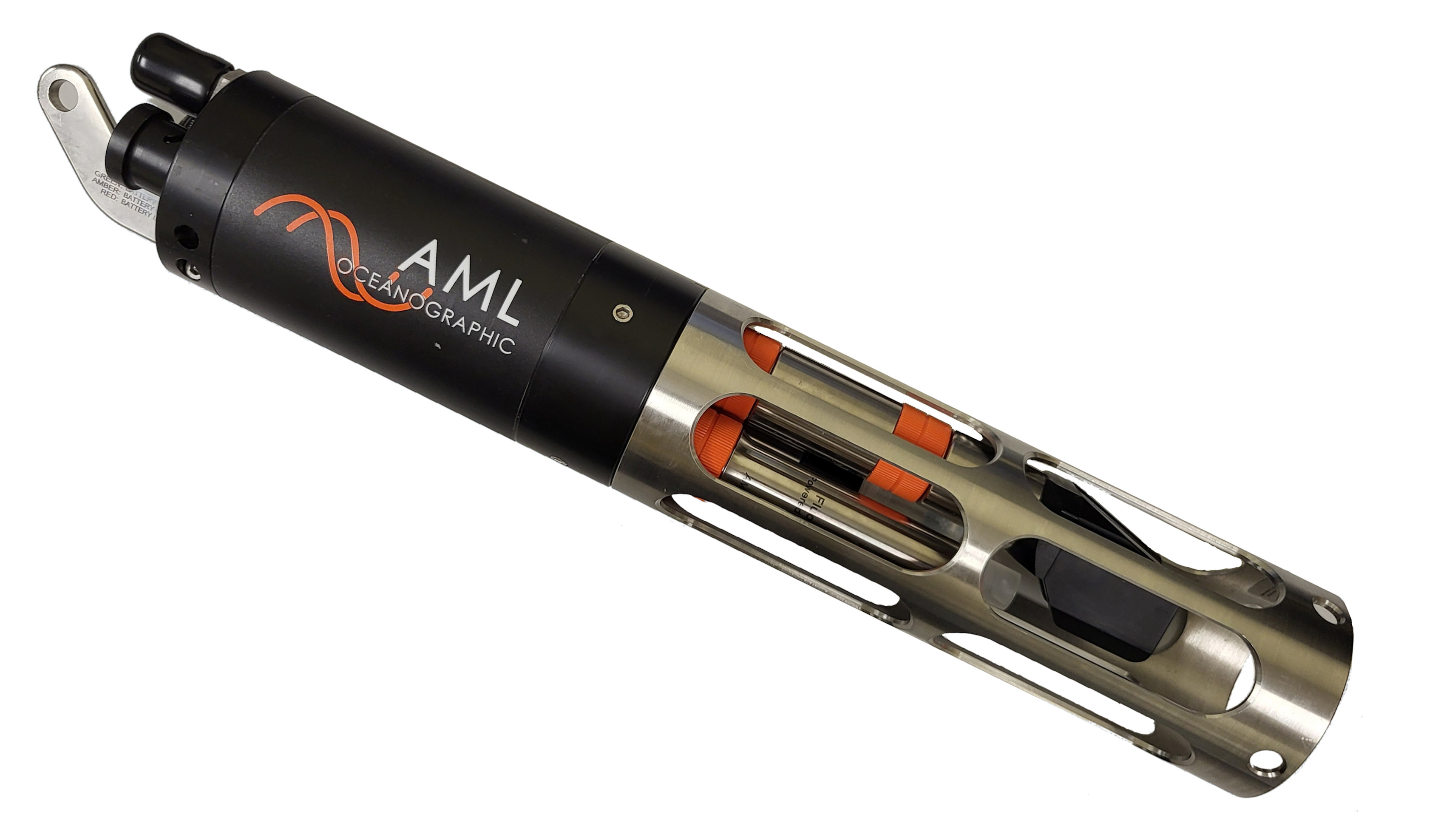}
        \caption{}
        \label{subfig:aml_sensor}
    \end{subfigure}
    \begin{subfigure}{0.45\linewidth}
        \includegraphics[width=0.9\linewidth]{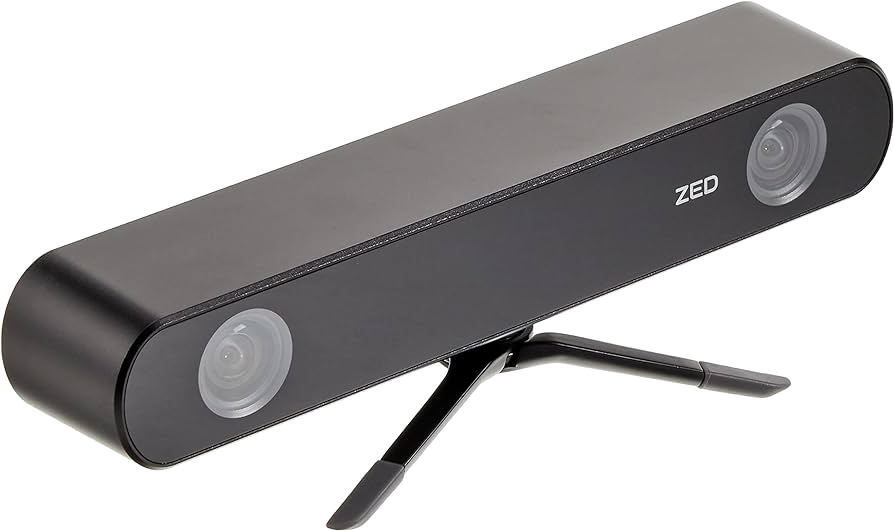}
        \caption{}
        \label{subfig:zed2}
    \end{subfigure}
    
    \begin{subfigure}{\linewidth}
        \centering
        \includegraphics[width=0.5\linewidth]{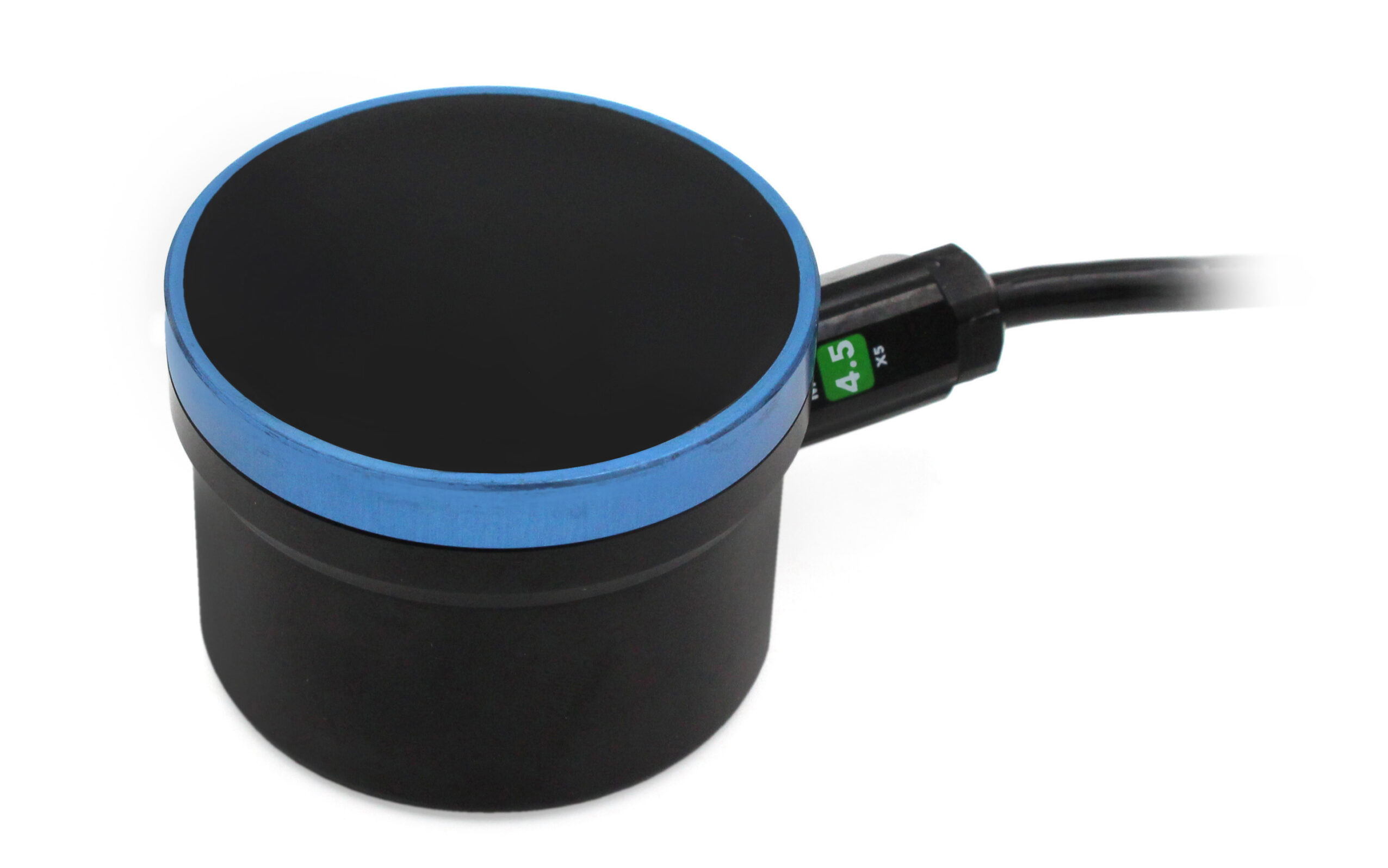}
        \caption{}
        \label{subfig:sonar}
    \end{subfigure}
    \caption{The three sensors available in the prototype. In (a), the AML WQP probe. In (b), the ZED2i camera. In (c), the BlueRobotics Ping1 Sonar.}
\end{figure}
The proposed ASV is equipped with several components that enable its autonomous navigation and data collection in aquatic environments. 
The Jetson Xavier NX\footnote{https://www.nvidia.com/es-es/autonomous-machines/embedded-systems/jetson-xavier-nx/} is an embedded computing platform designed for AI-driven applications.
It serves as the central processing unit for the ASV and can integrate data from various sensors, such as cameras and sonars among others, to perform sensor fusion. Its GPU-accelerated architecture enhances the ASV's perception capabilities using AI-based techniques,  allowing it to detect obstacles, optimize path planning and make real-time decision-making for autonomous navigation.

In addition to the Jetson Xavier NX, the ASV incorporates the Navio2\footnote{https://navio2.hipi.io/} autopilot, which operates as a HAT connected to a Raspberry Pi 4. Running Ardupilot\footnote{https://ardupilot.org/} inside, this autopilot is responsible for providing additional sensors, interfaces and processing power dedicated to navigation and autonomous control tasks, such as thruster actuation and waypoint routing algorithms. Communication between the devices is facilitated through a local network, with each device connected via an Ethernet cable to an industrial modem which provides internet connection via a IoT 4G SIM card. Additionally, this modem serves as a WiFi hotspot, enabling remote connectivity to the ASV. This configuration allows external control and monitoring of the devices, providing flexibility and accessibility in the ASV operating environment.

To achieve centimeter accuracy, a GPS/GNSS RTK system is implemented. The Emlid Reach M+ module with a Tallysman antenna is located on the vehicle, connected directly to Navio2 to provide positioning information, and communicates via LoRa\footnote{https://lora-alliance.org/} with the Emlid Reach RS+ ground station, which sends real-time corrections. This allows the vehicle to be handled in narrow areas by having it always well positioned with such accuracy. The movement of the vehicle is carried out by two BlueRobotics T200 thrusters, located at the rear of the catamaran. These are compact but powerful brushless direct current motors (BLDC) controlled by pulse width modulation (PWM) signals. They are specially designed for ROVs, AUVs and surface vessels, among others, thus providing agility and precision in vehicle handling. 

\begin{figure}[t]
    \centering
    \includegraphics[width=1\linewidth]{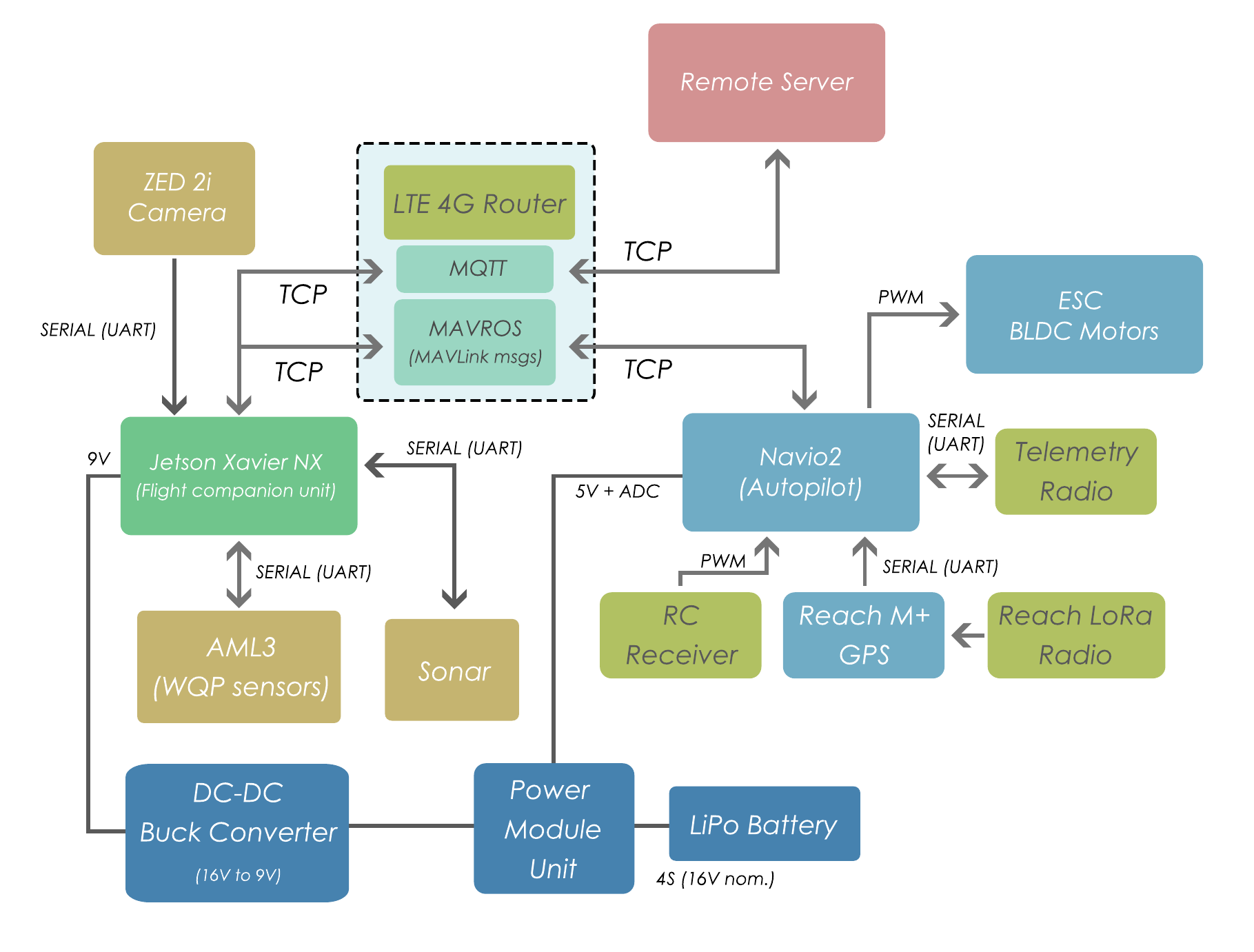}
    \caption{Architecture of ASV hardware components.}
    \label{fig:hardware_arch}
\end{figure}

The Ping2\footnote{https://bluerobotics.com/store/sonars/echosounders/ping-sonar-r2-rp/} sonar (see Fig. ~\ref{subfig:sonar}) from Blue Robotics offers the ASV underwater depth sensing capabilities. The device is a single-beam echosounder that can measure distances underwater, reaching depths of up to 100 meters. 
It is connected to the Jetson Xavier NX through a UART to USB adapter. Moreover, data retrieval and processing is done using the Python library provided by Blue Robotics. The sonar allows the ASV to accurately map the underwater terrain and depth contours of water bodies, making it useful for bathymetry measurements, which are fundamental to conduct an appropriated monitoring of water resources. Additionally, it can be used as an underwater obstacle avoidance sonar to navigate safely in challenging environments. 

The Stereolabs Zed 2i\footnote{https://www.stereolabs.com/products/zed-2} (see Fig. ~\ref{subfig:zed2}) is a stereo camera system that enhances spatial awareness by providing depth perception and visual data for various applications. It is connected to the Jetson Xavier NX through USB 3.1, and the ZED API provides low-level access to the camera and sensors. The ASV can perform object detection and localization tasks using advanced computer vision algorithms deployed on the Jetson Xavier NX, leveraging the depth information provided by the ZED 2i camera.

All of the above on-board ASV devices, including propulsion, computing and sensor systems, are powered by a set of two LiPo batteries in parallel, with a nominal voltage of 14.8V and 10.000 mAh each one. The current configuration of 20.000 mAh provides an autonomy of about 2 hours, but can easily be increased by doubling or tripling the capacity by simply adding more batteries in parallel, ensuring reliable and uninterrupted operation during extended missions. 

\begin{figure}[t]
    \centering
    \includegraphics[width=1\linewidth]{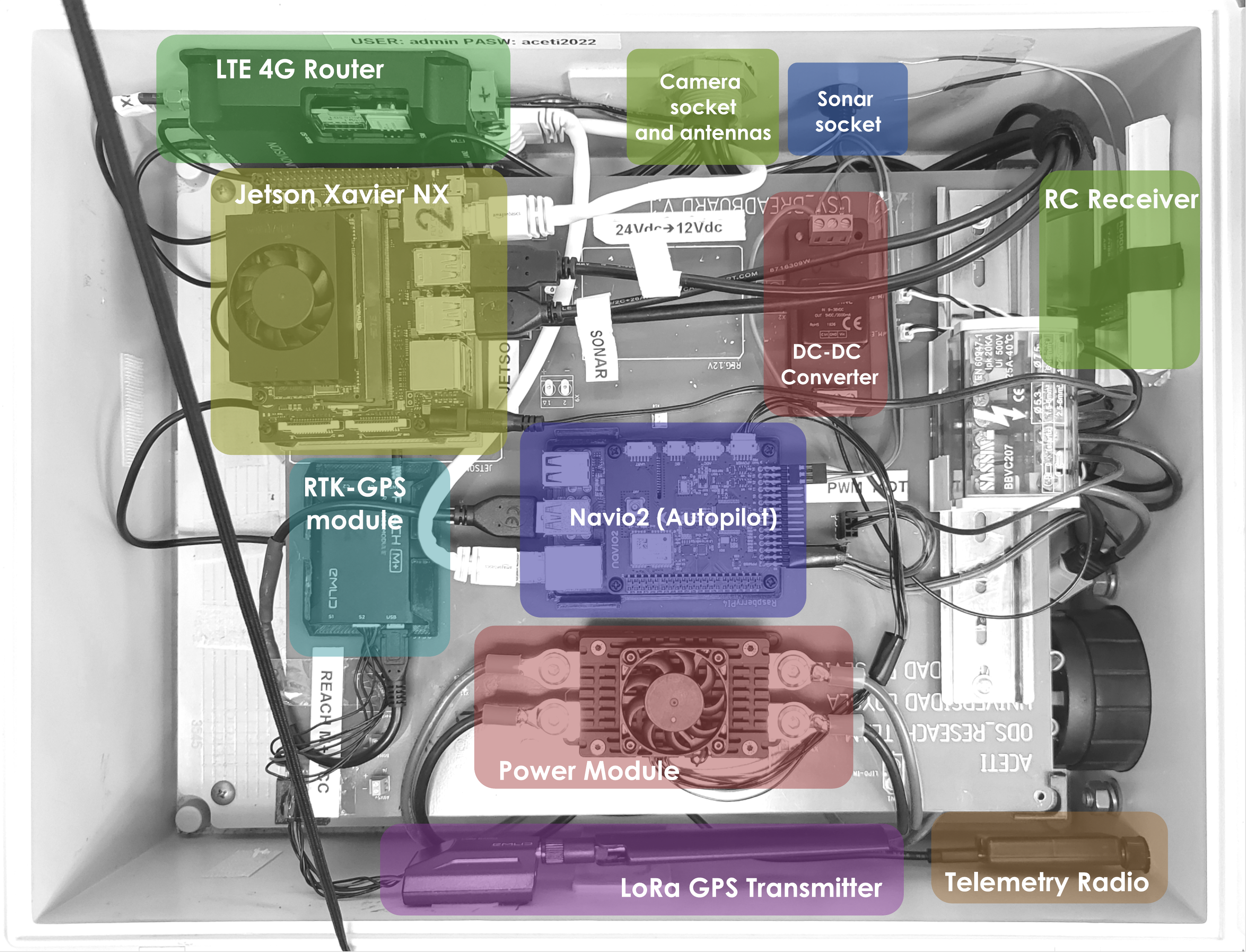}
     \caption{Photo of the hardware components inside of the ASV prototype.}
    \label{fig:hardware_photo}
\end{figure}

The ASV is equipped with the AML-3 XC Oceanographic\footnote{https://amloceanographic.com/aml-3-flexible-oceanographic-instrument} multi-sensor (see Fig. ~\ref{subfig:aml_sensor}) for measuring water quality parameters. This instrument has three sensor ports that allow for modular sensor interchangeability, making it adaptable to specific project needs. Certain sensors, such as Conductivity Temperature ones, offer dual-parameter measurement, which enables simultaneous tracking of up to four distinct parameters. The AML-3 measures pH, conductivity, temperature, and turbidity, providing crucial data on water quality and environmental conditions. Additionally, it supports vertical profiling of the water column, with a maximum depth capability of 500 meters. Communication with the Jetson Xavier NX is facilitated through an RS232 to USB cable, using command-based protocols for data retrieval. The sensor is self-powered by an internal rechargeable battery and features a mechanical on/off switch for control.
The hardware architecture of the ASV is shown in Fig. ~\ref{fig:hardware_arch}, which illustrates the arrangement of components and their connections. Fig. ~\ref{fig:hardware_photo} provides a visual representation of all components placed within the ASV.

\section{Software architecture}
\label{Software architecture}

The core of the software is divided in three main pieces of hardware: First, the aforementioned Xavier NX will serve as a Flight Companion Board (FCB) to manage the high level behavior of the vehicle, i.e., the goal waypoint query, the flight modes, the vision subsystem, etc. Secondly, the Navio2 is in charge of the low-level navigation system with Ardupilot \footnote{\url{https://ardupilot.org}} as the main middleware. Ardupilot is an open-source autopilot software that can handle multiple types of unmanned vehicles and serves as a low-level controller with a standard communication interface called MAVLink \footnote{\url{https://mavlink.io/}}. Finally, the last computation unit is located outside of the ASV, as a remote server within the computation infrastructure of the University of Sevilla. This server, equipped with multiple high-end GPUs and parallel computing CPU capabilities, is used as a central computation unit for the ASVs, to produce water contamination models based on machine learning, like Gaussian Process models, which computation complexity is too high to be handled locally. Therefore, the AI-based movement policy of the ASV is implemented in that server. This feature is suitable for battery-supplied ASVs that the one proposed in this paper since it alleviate the computational charge of the system.

\begin{figure}[tbp]
    \centering
    \includegraphics[width=1\linewidth]{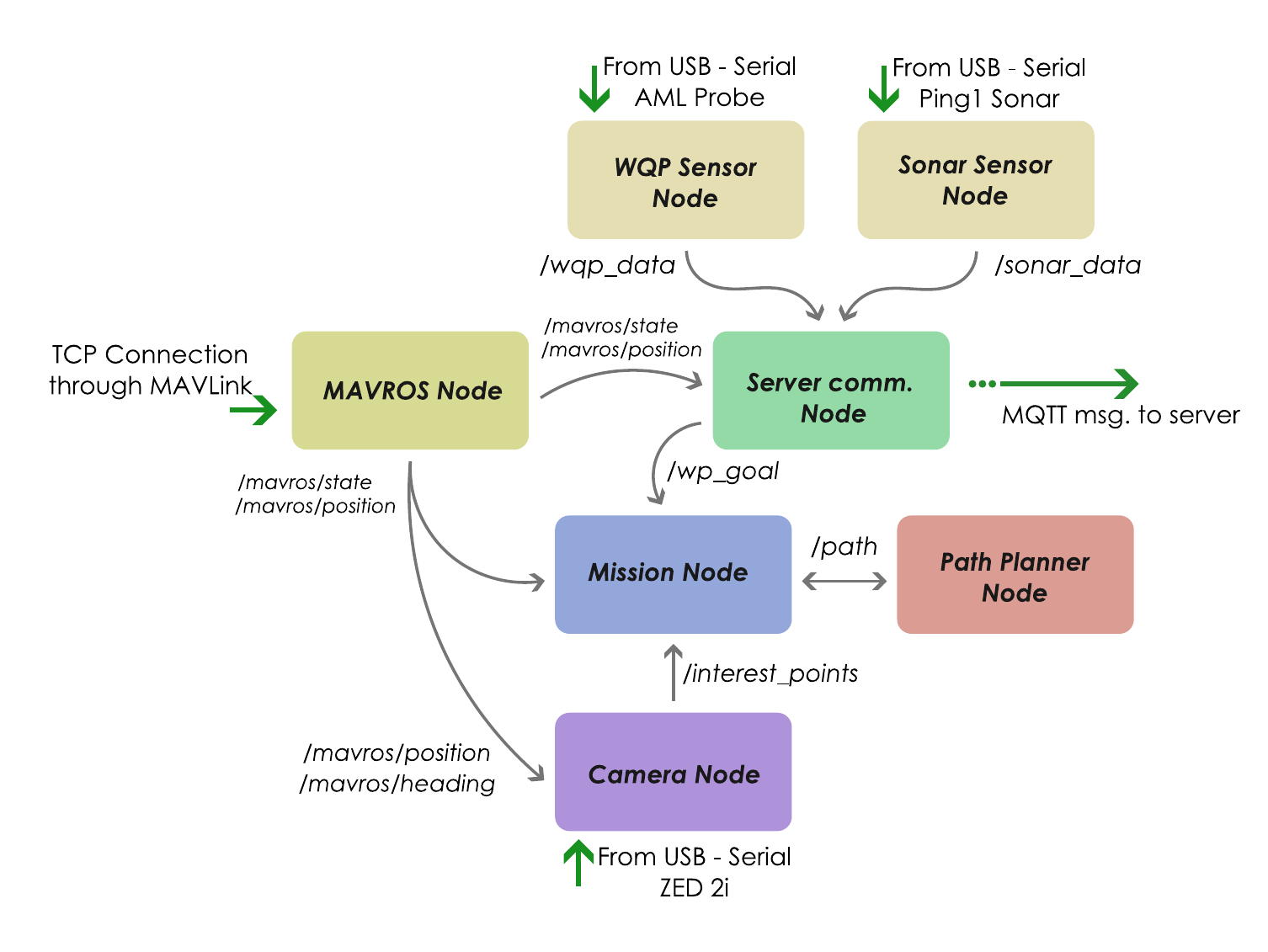}
    \caption{ROS2 software dockerized for easy deployment. The green arrows corresponds to the external interfaces to other peripherals.}
    \label{fig:ros2}
\end{figure}

The FCB, with an Ubuntu 20.04 LTS, implements the software using ROS2 with Humble Hawskbill version, an open source robotic middleware that provides a comprehensive set of libraries and tools designed for developing robotic systems. Leveraging the extensive capabilities of ROS2, additional sensors, actuators and peripherals can be integrated into the ASV operating framework. within this framework, the several messages and data structures, like RGB images, waypoint queries, and so on, are published in topics. These topics can be subscribed by a ROS2 node within the software, which enhances the flexibility and implementation of new components of the architecture. ROS2 provides also with common interfaces, which helps defining custom nodes and community nodes following the principle of \textit{not reinventing the wheel}. The different topics and nodes are depicted in Fig. \ref{fig:ros2}. The main implemented nodes in the proposed ASV, are:

\begin{itemize}
    \item \textbf{Mission Node}: In charge of receiving the goal WP and points of interest, i.e., points of water contamination, sanitize them, and inject them into the Ardupilot via MAVLink through the MAVROS node. 
    \item \textbf{Path Planner Node}: This node implements a Dijkstra path planner \cite{dijkstra1959note} that optimally obtains an off-line feasible route from the current ASV position to the goal WP.
    \item \textbf{WQP/Sonar Node}: These nodes handle the serial communications of the FCB with the WQP sensor and the sonar. They translate the dataframes to ROS2 messages for a given node to subscribe.
    \item \textbf{Camera Node}: Node in charge of implementing the AI YOLO modules for obstacles and macro-plastic detection. This node provides the Mission Node and the server with goal WPs.
    \item \textbf{Server comm. Node}: This node handles the bi-directional communication with the central server. 
    \item \textbf{MAVROS Node}: Handles the communication with Ardupilot via a MAVLink standard interface.
\end{itemize}

\begin{figure}[tbp]
    \centering
    \includegraphics[width=1\linewidth]{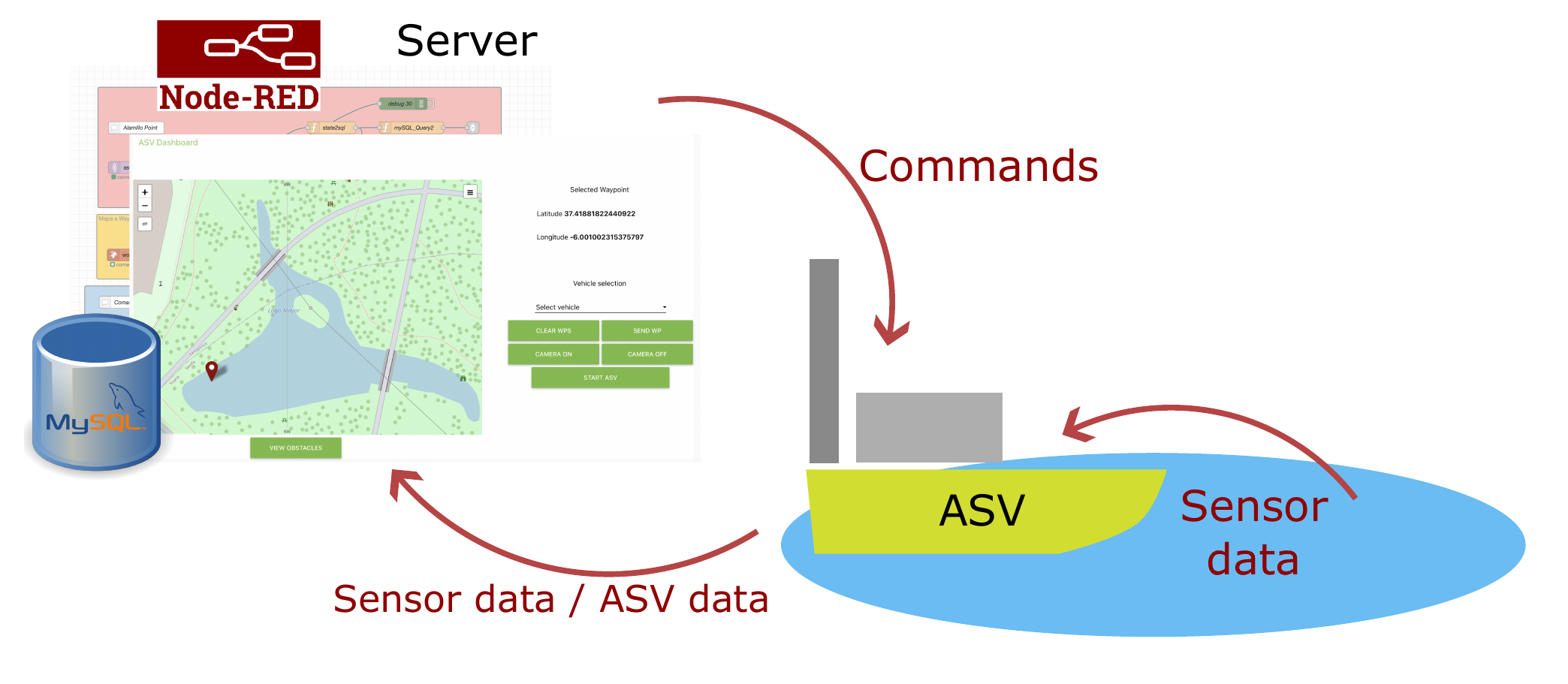}
    \caption{Schematic of the communication used in the ASV.}
    \label{fig:water_to_server}
\end{figure}

This proposed ROS2-based software is embedded in a Docker\footnote{\url{https://www.docker.com}} image to support cross compatibility and to abstract the software from the hardware, and to be tested in different AMD/ARM architectures.

The central server is an Ubuntu 22.04 LTS server with 2x GPUs (Nvidia RTX 3090 and Nvidia RTX A4000) and dual Intel Xeon CPU architecture. This server will serve as an proxy-computation server for the vehicle. The server communicates with the ASV through the Internet using Message Queuing Telemetry Transport\footnote{https://mqtt.org/} protocol (MQTT). This lightweight IoT protocol, enables a fast and reliable communications within nodes. One of the advantages of this protocol is the small size of the dataframes, which is important when the bandwidth is small and the data consumption of the IoT 4G SIM inside of the ASV is limited. The central server is also in charge of providing a Graphical User Interface (GUI) for the supervision of the ASV. This GUI, implemented using Node-Red \footnote{https://nodered.org/}, serves a double purpose: First, it filters and manages the operative information that the ASVs transmits (position, speeds, safety flags, etc). Second, it can be used to operate the ASVs remotely using an Informative Path Planner \cite{yanes_deep_2020} implemented inside of the server, in a rendezvous architecture. 

The server will also receive the data from the sensors to simultaneously conform a water quality models based on machine learning algorithms, as it will be explained later. Thus, the high-end hardware capabilities of this server leverages the computing skills of the ASVs as a on-proxy GPU, for example, when Deep Learning is used for Informative Path Planning algorithms like in \cite{yanes_deep_2020}. This server is also used as a mySQL\footnote{https://www.mysql.com/} database for the water quality models and data generated. Thus, a robust and scalable relational database management system is implemented, which allows to have a structured repository for storing and organizing research mission data.

\section{Experimental Results}
\label{Experimental Results}

The tests carried out have successfully validated the autonomy, sensor accuracy, and communications capabilities of the proposed ASV, paving the way for future exploration and monitoring missions in aquatic environments. The tests were carried out in Lago Mayor, Parque del Alamillo (Sevilla), a relatively reduced artificial lake with a large number of flora and fauna. Several field tests were conducted to evaluate the autonomy and data collection capabilities of the ASV. Using two 4-cell lithium batteries, the ASV achieved an autonomy of around 2 hours while maintaining an average velocity of 1 m/s. This demonstrates its ability to operate autonomously for extended periods without the need for recharging.

\begin{figure}[tbp]
    \centering
    \includegraphics[width=0.9\linewidth]{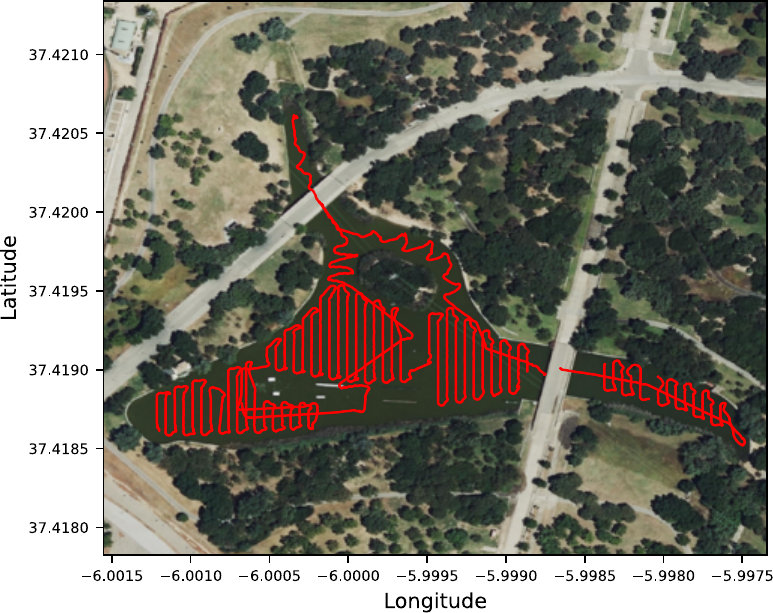}
    \caption{Path followed by the ASV during the monitoring mission of Lago Mayor, in Parque del Alamillo (Sevilla).}
    \label{fig:path_alamillo}
\end{figure}

During the course of the mission of a length of about 3 km (as can be seen in Fig. \ref{fig:path_alamillo}), it was demonstrated the ability of the ASV to execute planned missions using a series of waypoints, as well as its ability to maintain stable communications with the server to transmit the data captured by the sensors in real time while operating on the ground.

\subsection{Water Quality Parameters and Bathymetry estimation}

\begin{figure}[tbp]
    \centering
    \includegraphics[width=1\linewidth]{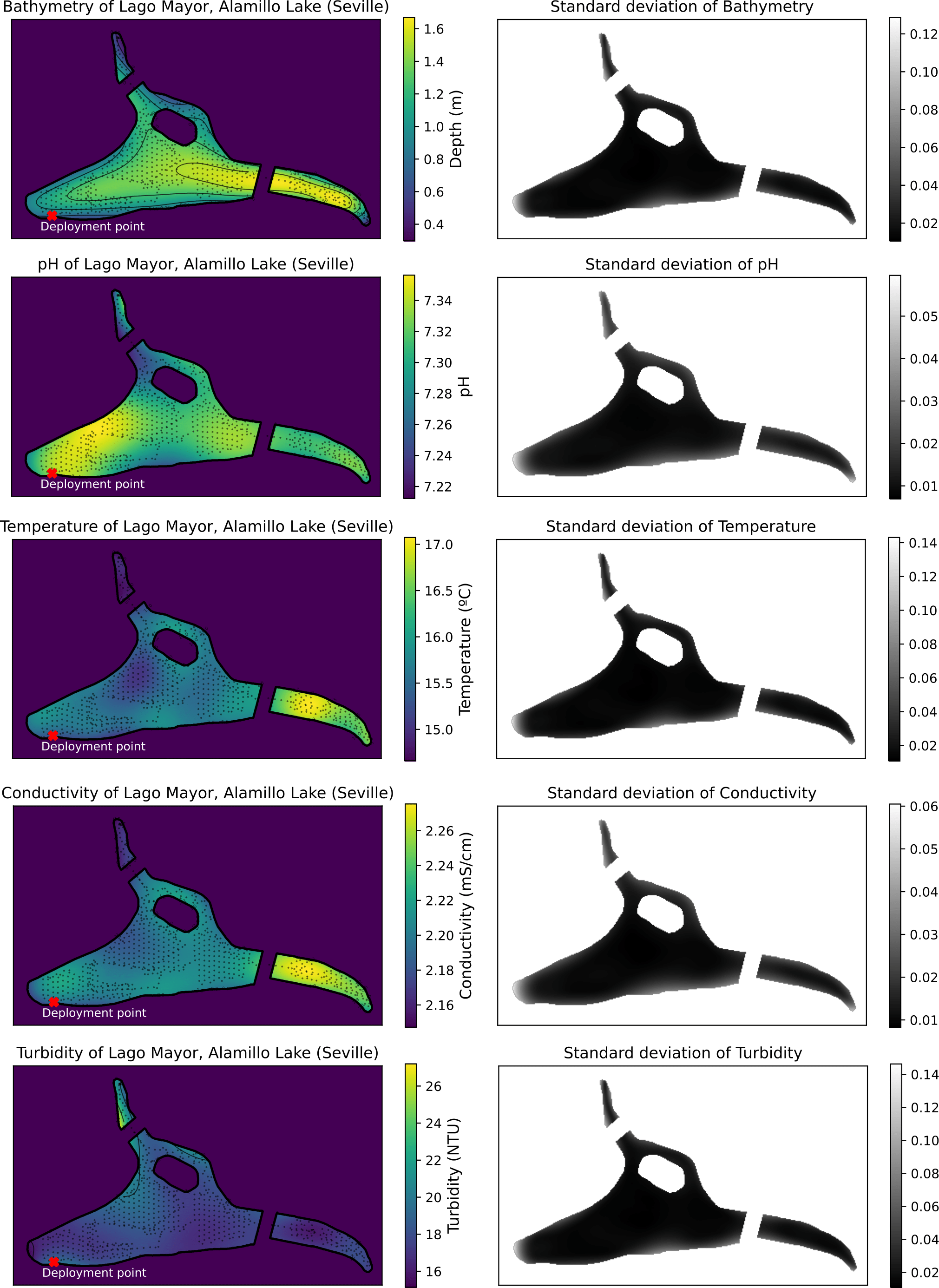}
    \caption{Bathymetry and WQP estimations generated from the Lago Mayor monitoring mission during winter season.}
    \label{fig:results_alamillo}
\end{figure}

Data collected during the monitoring missions include bathymetry measurements, pH, temperature, conductivity, and turbidity readings of the lake. These data provide detailed information about the aquatic environment, crucial for understanding the ecological status of the waterbody and identifying any potential environmental concerns. The parameter maps depicted in Fig. \ref{fig:results_alamillo} have been generated using Gaussian processes (GPs), a machine learning modeling technique commonly employed in spatial data analysis. As seen in \cite{rasmussen_gaussian_2006}, by representing functions as random variables with Gaussian distributions, GPs offer a principled framework for incorporating prior knowledge and making predictions based on observed data. Thus, they are particularly well-suited for interpolating and extrapolating data points in continuous spatial domains, such as those encountered in environmental monitoring missions. The kernel used by the implemented GP is the one known as Radial
Basis Function (RBF), especially useful in environments where a smooth and continuous variation of parameters is expected, as seen in \cite{samaniego2021bayesian}. This kernel is defined by a lenghtscale parameter, which determines the distance between correlated samples, which has been initialized in this case at around 80 meters, with approximate bounds between 55 and 110 meters. 

The findings obtained during the study were compared with the international standards \cite{world1963international} and with the Spanish regulations \cite{realdecreto2023} for drinking water, and it was simply verified that the water was not suitable for human consumption, as expected in this type of lake, with the most undesirable parameter being the high turbidity.

\subsection{Macro-plastic detection and localization}

\begin{figure}
    \centering
    \includegraphics[width=1\linewidth]{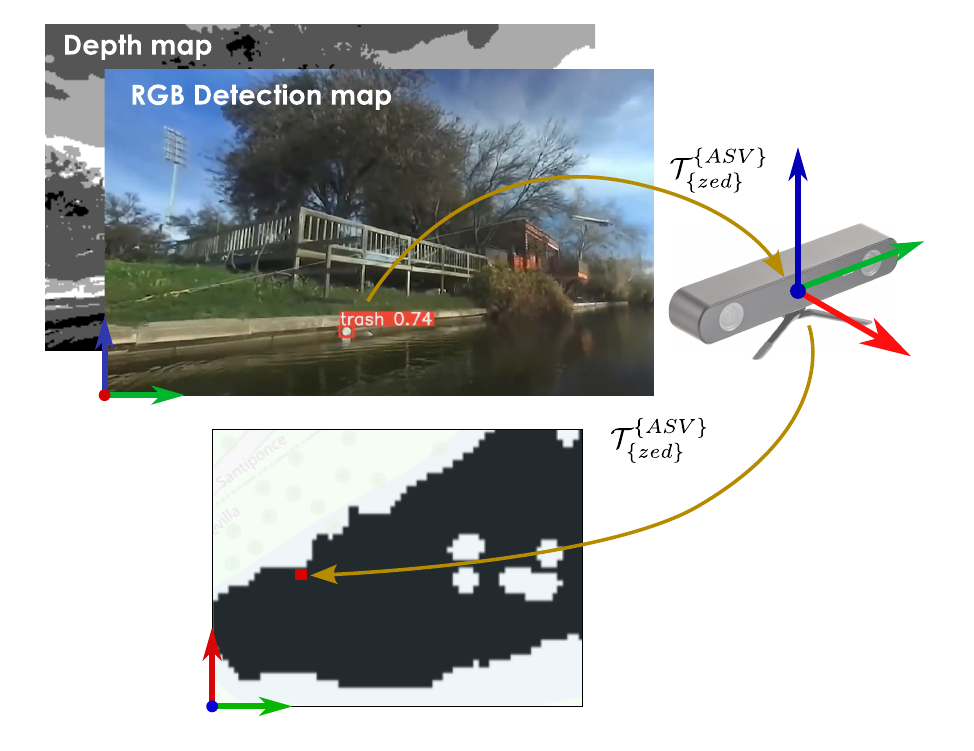}
    \caption{Depth map + YOLOv5 detection scheme of macro-plastic in a real scenario in Lago Mayor, Parque del Alamillo, Sevilla. In red, the bounding box of the detection with its predictive confidence. Using the GPS positioning and attitude of the vehicle, is possible to locate the macro-plastics in a global frame.}
    \label{fig:zed_yolo_scheme}
\end{figure}

The ZED2i stereo camera captures both RGB and depth images simultaneously, providing valuable visual and spatial information about the ASV's surroundings. 
The RGB image obtained from the left eye of the ZED2i camera serves as input to a YOLOv5\cite{kim2022object} custom model running on the powerful GPU of the Jetson Xavier NX. The YOLOv5 algorithm is a highly efficient and accurate method for object detection. It has been fine-tuned to perform real-time detection of floating debris and pollutants, such as macro-plastics on water surfaces (see Fig. ~\ref{fig:zed_yolo_scheme}).
The ZED2i stereo system offers accurate depth information for each pixel in the RGB image. By analyzing the depth data of detected objects, their distance from the camera in meters is precisely estimated, and their positioning with respect to the ASV reference frame is determined. This information is combined with the ASV's GPS coordinates to locate macro-plastics accurately in a global frame of reference (see Fig. ~\ref{fig:zed_yolo_scheme}).
The ASV transmits the detected macro-plastics and their global coordinates to the mission node as interest points. This information enables the ASV to contribute to the preservation of aquatic ecosystems and the mitigation of water pollution by providing targeted cleanup areas.

\section{Conclusion and Future Works}
\label{Conclusion and Future Works}

An innovative ASV has been introduced for aquatic exploration and monitoring, designed to navigate through extensive body waters while gathering a wide range of environmental information through the integration of onboard sensors such as water quality parameters, sonar and camera. 
In future missions, the deployment of a fleet of ASVs, capable of real-time cooperation and task distribution, is considered. ASVs can be assigned multiple tasks simultaneously, such as intensification, which consists of prioritizing areas based on contamination level, or exploration, which involves uniformly covering the water body. 
Cooperative missions may also involve fleets of heterogeneous vehicles with different measurement or movement capabilities.
Future research will implement Deep Reinforcement Learning (DRL) algorithms, such as those from \cite{yanes2023deep}, to enable Informative Path Planning as the samples are taken from the lake.
Furthermore, techniques from \cite{luis2024deep} will be incorporated by combining Local GPs and DRL to optimize monitoring policies. This will include the incorporation of collision-free Consensus-based heuristics for the safe deployment of multiple surface vehicles, as well as a unified neural network for all agents.
In addition, the ASV design will be modified to integrate a winch for submerging AML Oceanographic sensors and collecting samples at various depths, expanding monitoring capabilities. This enhancement enables vertical profiling of the water column, offering valuable insights into water quality dynamics.

\bibliography{refs}
\bibliographystyle{IEEEtran}

\end{document}